# DIRI: Adversarial Patient Reidentification with Large Language Models for Evaluating Clinical Text Anonymization

*John X. Morris[1]\*, Thomas R. Campion, Jr., PhD[2]; Sri Laasya Nutheti[1]*
*Yifan Peng, PhD[2]; Akhil Raj[1]; Ramin Zabih, PhD[1]; Curtis L. Cole, MD[2]\**
*[1] Cornell Tech, New York, NY   [2] Weill Cornell Medicine, New York, NY*

**Abstract**

*Sharing protected health information (PHI) is critical for furthering biomedical research. Before data can be distributed, practitioners often perform deidentification to remove any PHI contained in the text. Contemporary deidentification methods are evaluated on highly saturated datasets (tools achieve near-perfect accuracy) which may not reflect the full variability or complexity of real-world clinical text and annotating them is resource intensive, which is a barrier to real-world applications. To address this gap, we developed an adversarial approach using a large language model (LLM) to re-identify the patient corresponding to a redacted clinical note and evaluated the performance with a novel De-Identification/Re-Identification (DIRI) method. Our method uses a large language model to reidentify the patient corresponding to a redacted clinical note. We demonstrate our method on medical data from Weill Cornell Medicine anonymized with three deidentification tools: rule-based Philter and two deep-learning-based models, BiLSTM-CRF and ClinicalBERT. Although ClinicalBERT was the most effective, masking all identified PII, our tool still reidentified 9% of clinical notes Our study highlights significant weaknesses in current deidentification technologies while providing a tool for iterative development and improvement.*

## 1. Introduction

The privacy of medical data is paramount for maintaining patient confidentiality and adhering to regulations such as the Health Insurance Portability and Accountability Act (HIPAA) [27]. Deidentifying medical data means removing or replacing personal identifiers so that an unauthorized party cannot determine a patient's identity[16]. Current deidentification tools are tested on datasets that annotate each type of protected health information (PHI), such as "name", "address", and "medical record number", separately, and consider successful obfuscation of all PHI types identified by HIPAA to correspond with perfect performance.

When all personal identifiers are perfectly removed from a clinical note, is it still possible to identify a patient? To answer this question, we propose a new method, DIRI (De-Identification/Re-Identification) for evaluating the privacy of deidentified data, based on a system that uses large language models to reidentify patients from redacted clinical notes. Our method relies on the principle that a note is deidentified if our adversarial neural reidentification system can no longer correctly identify the patient from the note. This is analogous to "encryption" which is always relative to the current level of computational capability to decrypt.

Prior work has validated the effectiveness of "linkage attacks", which combine partial pieces of information to identify individuals from partially redacted datasets. For example, Sweeney[15] has shown that even though neither gender, birth dates nor postal codes uniquely identify an individual, the combination of all three is sufficient to identify 87% of individuals in the United States. In our experiments, we observe that our system naturally learns to combine quasi-identifiers into successful linkage attacks. These quasi-identifiers are not adequately addressed by traditional supervised methods, presenting a significant gap in existing deidentification techniques [4, 9, 10, 11, 12].

In this work, we aim to address this gap by targeting a more general definition of PHI. Inspired by the concept of K-anonymity [12], our approach involves learning a probabilistic reidentification model to predict the true identity of a given text. To validate our idea, we compare the performance of various deidentification tools on a dataset of real medical notes from Weill Cornell Medicine (WCM). We consider the performance of these deidentification tools across two axes: the amount of information redacted, and the performance of our reidentification system when taking the deidentified notes as input. This study evaluates the effectiveness of various masking tools in deidentifying a private dataset of patient demographics and EHR notes. We evaluate reidentification accuracy across varying levels of text masking for each tool, allowing us to assess the balance between privacy protection and information retention.

Notably, even with all PHI identified by current anonymization methods masked, our system is still able to correctly reidentify 9% of patients in a database of several thousand. By analyzing the reidentification rates across various deidentification strategies, we gain insight into how well each tool strikes a balance between safeguarding patient privacy and preserving critical medical information for downstream analysis. We hope that our new method can inspire future work in developing better tools for both deidentification and reidentification.

---

\* *Apart from first and senior author, authors are ordered alphabetically by last name.*

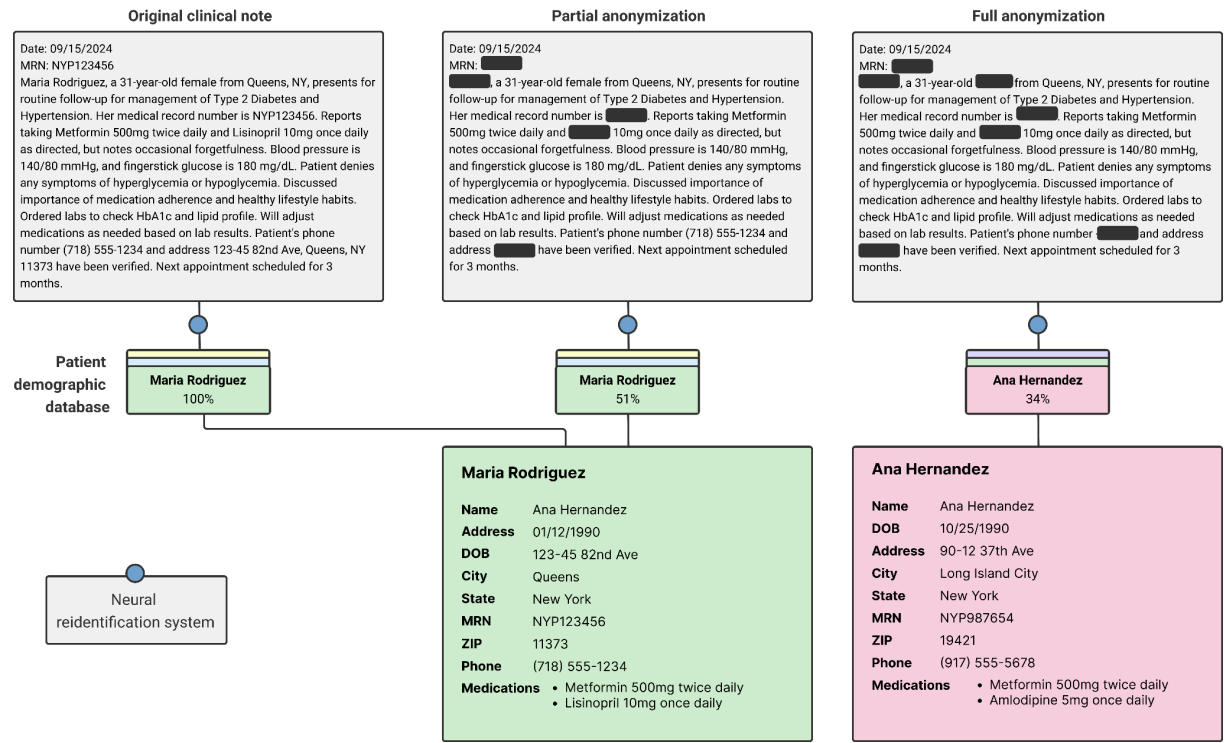

**Figure 1:** Our neural reidentification system identifies patients in a database from redacted clinical notes. After redacting a significant amount of information (which may be PHI) from the clinical note, our system can no longer identify the correct patient.

## 2. Related Work

**Deidentification datasets and evaluation.** A few datasets exist for evaluating deidentification systems, including the i2b2 deidentification datasets[14], the MIMIC deidentification dataset[9], and the Text Anonymization Benchmark[8]. Since deidentified datasets naturally contain sensitive information, these datasets either require credential verification for access, or describe non-medically-relevant data that is not technically private. These datasets typically evaluate deidentification methods using some combination of precision and recall on labeled tokens such as $F_1$ or $F_2$ scores. Our method is general (can be applied to any deidentified dataset) and does not require any human-labeled data.

**Automated deidentification.** The task of deidentification has been approached using several methods, such as neural networks[19], regular expressions[1], and Conditional Random Fields[20], and replacing PHI with surrogate values[20]. Broadly, they can be divided into two types: rule-based and machine learning-based (ML-based). Rule-based deidentification methods offer several benefits such as minimal or no model-training requirement, quicker runtime, high recall value[21], and usability on any free-text medical record[22], but do not perform as well as ML-based models in terms of finding quasi-identifiers since quasi-identifiers are difficult to manually find and so it is difficult to manually create a rule for them[3]. ML-based models, on the other hand, can utilize the complex trained features of their architecture to identify quasi-identifiers. Adversaries can use the presence of quasi-identifiers to perform reidentifications. Moreover, supervised approaches cannot naturally detect quasi-identifiers, since these words are not inherently labeled as PHI[18].

**Deidentification evaluation.** Data-sharing initiatives such as the All of Us project have used adversarial reidentification techniques to determine reidentification risk of non-text data[29]. Several previous studies have compared rule-based models and ML-based models at deidentifying text. Specifically, El-Hayek et al (2023) showed that rule-based tools perform better than ML-based tools and especially Philter provides high recall and flexibility, but it requires extensive revising of its pattern matching rules and dictionaries to reduce the false positives and so achieve high precision[21]. Pépin et al (2022) showed that TiDE[28] (Text DEidentification) utilizes a hybrid of rule-based and ML methods to achieve deidentification, but performs worse against a rule-based model in terms of speed and accuracy[20]. Furthermore, ML-based models are difficult to train if there is infrequent presence of identifiers in the corpus[19]. But, with sufficient amounts of data and effort put into training, ML models can eventually perform better than rule-based tools[19]. We aim to further the work of comparison of deidentification tools by comparing one rule-based (Philter) and two ML-based tools (BiLSTM-CRF and ClinicalBERT) using our novel method.

## 3. Materials and Methods

In this section, we first describe the generic template for our reidentification model, which learns to identify patients from clinical notes anonymized using a certain deidentification tool. We then describe a case study using medical records from Weill Cornell Medicine including data collection, processing, and the settings of particular deidentification tools.

### 3.1 Reidentification Model

We built a neural reidentification model to identify patients in a database from partially redacted text. To validate a given deidentification tool, we first deidentified a large number of patient notes and separated them into "train" and "test" notes. We then trained our reidentification model on the deidentified training set and measured its performance on the deidentified testing set. In our framework, a perfect deidentification will result in 0% top-1 test accuracy.

We first built a reidentification model (Figure 1) to identify the patient described in a partially deidentified patient note. Suppose we have a large database of clinical notes of $X$ and corresponding patient metadata or "profiles" of $Y$. Our goal is to learn to identify the correct patient $y$ given redacted clinical note $x$. We can formulate this probabilistically as p(y | x; θ), where θ represents the parameters of a neural network. We then train θ to maximize the likelihood p(y | x; θ) using standard backpropagation.

One way to parameterize $p(y|x; \theta)$ would be to compute this value pairwise for all possible profiles and documents. However, this computation scales with $O(mn)$ for m documents and n profiles; running this number of computations is intractable for a large number of notes or patients, which we expect in a realistic clinical setting. To be able to efficiently compute p for an entire database, we decomposed this function into representations of $y$ and $x$ to be computed separately:

$$p(y|x; \theta) = softmax_x[f(y; \theta) \cdot g(x; \theta)]$$

Here, $f$ and $g$ are separate networks that produce vector-wise representations of documents and profiles, compared using the simple dot product $f(y) \cdot g(x)$. This setup is typical of a "biencoder model" across many fields including text retrieval[24] and recommendation systems[25]. To train this model end-to-end, we alternately optimize $f$ and $g$ using a "coordinate ascent" strategy to maximize the likelihood of a patient profile $y$ given a redacted profile $x$. For further details see Morris et al (2022)[3].

### 3.2 Deidentification Tools

For this study, we selected three deidentification tools—Philter[1], BiLSTM-CRF[2], and ClinicalBERT[13]—based on their diverse methodologies and relevance in the field of medical data privacy (Table 1). These tools were chosen for their open-source availability, popularity, and unique strengths. Philter, is a widely-used rule-based tool known for its effectiveness in identifying and masking PHI using pattern recognition and customizable rules, making it a strong baseline for rule-based approaches. The BiLSTM-CRF tool was developed for Dutch medical records, using BiLSTM-CRF, a deep learning method that combines statistical models and neural networks. Lastly, ClinicalBERT[26] is a cutting-edge ML-based tool using BERT models trained on clinical textual data, representing the latest in neural deidentification techniques. These tools allowed us to assess the full spectrum of deidentification methods, from rule-based to advanced NLP, ensuring comprehensive evaluation in this study.

**Table 1:** Comparative Summary of PHI deidentification tools.

| Tool | Overview | Key Features and Methods |
|---|---|---|
| Philter[1] | Combines rule-based and statistical NLP methods to remove PHI while retaining non-PHI. | **Pattern Matching**: Uses regex for identifying PHI patterns (dates, names, etc.). <br> **Part-of-Speech Tagging**: Identifies proper nouns as potential PHI. <br> **Blocklist**: Known PHI terms are marked. <br> **Allowlist**: Reduces false positives by excluding common non-PHI terms. <br> **Safety Net**: Any token not explicitly safe is reviewed. |

| | | **Final Filtering**: Captures remaining PHI after initial filtering.<br>**Masking:** Identified PHI is masked using asterisks. |
|---|---|---|
| ClinicalBERT[13] | BERT-based neural model fine-tuned for PHI deidentification in clinical text. | **BERT Model**: Uses pre-trained ClinicalBERT model for PHI tagging.<br>**BIO Tagging**: Identifies PHI using Beginning Inside Outside (BIO) scheme.<br>**Regex for Preprocessing**: Cleans input text before tokenization.<br>**Neural Architecture**: Leverages BERT's context understanding for high accuracy.<br>**Surrogate Data Generation**: Supports realistic surrogates for data utility, although masked words have been replaced with asterisks in this implementation to maintain consistency of outputs across the three masking tools. |
| BiLSTM-CRF[2] | BiLSTM-CRF model with pre-trained embeddings to identify and mask PHI. | **Tokenization & Segmentation**: Uses spaCy tokenizer and models.<br>**BIO Tagging**: Annotates text with BIO tagging scheme.<br>**Neural Model**: Combines BiLSTM with CRF output layer for high accuracy.<br>**PHI Identification**: Tags and masks PHI tokens.<br>**Post-Processing**: Ensures consistency in tagging.<br>**Surrogate Generation**: Replaces PHI with realistic surrogates, though masked words are replaced with asterisks in this implementation to maintain consistency of outputs across the three masking tools. |

Under each tool, we generated deidentified splits of our dataset by interpolating across various levels of masking. Both learned methods (BiLSTM-CRF and ClinicalBERT) yield a confidence score, allowing us to mask tokens to decrease the likelihood of being PHI. The rule-based Philter tool does not allow us to compute a confidence score, so we mask tokens labeled as PHI in a random order. In all cases, we replace each masked token with the special token "******".

### 3.3 Data collection

Weill Cornell Medicine (WCM) is an academic medical center in New York City with more than 1800 attending physicians conducting 3.09 million annual patient visits across more than 165 locations. WCM physicians hold admitting privileges at NewYork-Presbyterian Hospital. To document outpatient care, WCM physicians have used the EpicCare Ambulatory EHR system since 2000. The WCM Institutional Review Board approved this study. All data was reviewed and approved for use by the institutional review board (IRB) to ensure compliance with ethical standards and patient confidentiality.

The dataset[31] provided for this study by WCM consisted of patient demographics and corresponding EHR notes. The dataset contained clinical reports (including progress notes) in English from heart failure patients admitted and discharged at NewYork-Presbyterian/Weill Cornell Medical Center between January 2008 and July 2018. Patients were identified using billing codes ICD-9 Code 428 or ICD-10 Code I50, indicating a diagnosis of heart failure. The dataset was divided using a 70/15/15 ratio to obtain training, validation, and test datasets.

The original dataset (before our selection step) consists of 14.68 million reports(including thoracoabdominal computed tomography (CT) reports and progress notes) from 21,311 heart failure (HF) patients. Each report includes either a radiological finding or a progress note and their corresponding headers. An example of note headers is shown in Table 2. Each note header contains 14 identifiers.

**Table 2.** Note headers with examples ("morphed" data), HIPAA categories and counts.

| Category | Example | HIPAA? | Number of notes with identifier |
|---|---|---|---|
| Note datetime | 2013-12-06 00:00:00 | Yes | 16917 |
| Note class | Telephone encounter | No | 5467 |
| Medical Record Number (MRN) | 1234943 | Yes | 13083 |
| Gender | Female | No | 11170 |
| Date of birth | 03/30/1942 | Yes | 10653 |
| Race | White | No | 1793 |
| Ethnicity | Not Hispanic or Latino | No | 1126 |
| Death date | 2014-01-02 | Yes | 597 |

| Death datetime | 2014-01-02 03:30:00 | Yes | 597 |
| Address 1 | 412 ETHAN AVE | Yes | 7532 |
| Address 2 | APT 43D | Yes | 6945 |
| City | NEW YORK | Yes | 11727 |
| State | NY | No | 18059 |
| Zip | 10432-3243 | Yes | 6505 |

### 3.4 EHR Note Selection

To improve reidentification accuracy, we developed a custom EHR note selection method that focused on selecting one set of notes per patient, written by a designated physician, using patient ID as the key identifier. Initially, we selected the first EHR note (index-wise) for each patient, but this method often missed identifiable information, resulting in a reidentification accuracy of only 5%.

We then introduced the concept of *relevancy*, defined by the number of note headers present in the EHR note, increasing accuracy to 16%. To further improve performance, we increased the input's length from 128 to 256 tokens, allowing the model to process more information. Relevancy was recalculated using the first 256 tokens, which improved training accuracy to 28.5%. Finally, by increasing the input's length to 512 tokens, we further improved accuracy. Relevancy was recalculated based on the first 512 tokens, resulting in a final reidentification training accuracy of **90.28%**.

### 3.5 Neural network details

For *f*, the neural network that encoded EHR notes, we used the RoBERTa-base text encoder (Liu et al., 2019), which has 125 million parameters. For *g*, the neural network that encoded profiles, we employed TAPAS-base table encoder (Herzig et al., 2020), which has 111 million parameters. All models were implemented using the Hugging Face "transformers" library[30]. Each reidentification model was trained for 55 epochs with a batch size of 35. The maximum input sequence length was capped at 512 tokens, with longer documents truncated accordingly, ensuring that the model extracted the maximum amount of information within the architectural constraints. A learning rate of 1e-5 was applied using the Adam optimizer. The loss function utilized was coordinate ascent, which iteratively optimized reidentification probabilities to improve model accuracy. Each training session took approximately 10 hours on a single Quadro RTX 6000 GPU.

### 3.6 Evaluation

For each deidentified test split, we trained our profile and document encoders jointly on the corresponding deidentified train split. We then scored the performance of the deidentification tool using the top-k retrieval accuracy of our reidentification system. This metric is based on an anonymity criterion that requires the released information to be indistinguishable among at least a specified number of individuals, denoted by k. In our case, we set k = 1. This metric indicates how precisely the reidentification model can link EHR notes to patient demographics in the test set, inspired by the principle of k-anonymity[12].

### 4. Results

#### 4.1 Reidentification performance

Figure 2 plots reidentification accuracy (test) against the percentage of text masked for each tool tested. ClinicalBERT consistently outperformed both BiLSTM-CRF and Philter at all levels of masking. To better interpret the trends in reidentification accuracy, we categorized the masking levels into three ranges: low (0-5%), moderate (5-15%), and high (15-20%).

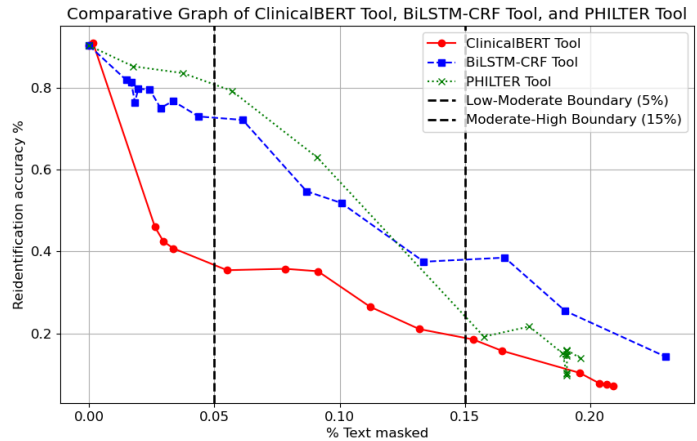

**Figure 2:** Reidentification test accuracy (y-axis) vs. the percentage of text masked by each tool (x-axis). Reidentification performance falls as the percentage of text masked increases.

**(i) Low Masking (0-5%)** In the low masking range, where minimal text is obfuscated, all tools maintain relatively high reidentification accuracy, which is inversely proportional to deidentification efficacy. Notably, ClinicalBERT demonstrates superior deidentification performance compared to the other tools, primarily due to its rapid reduction in reidentification accuracy. It starts with a high accuracy score of around 0.9, but experiences a sharp decline to approximately 0.4 as the masking increases to 5%. In contrast, BiLSTM-CRF and PHILTER exhibit a more gradual decline in accuracy over the same range, indicating a quite slow rate of improvement in the deidentification performance.

**(ii) Moderate Masking (5-15%)** In the moderate masking range, the performance of the tools diverges significantly. The ClinicalBERT tool demonstrates the best deidentification performance in this range. BiLSTM-CRF and Philter show a more consistent drop in accuracy(and so improvement in the deidentification performance) throughout the range, with Philter performing the weakest among the tools except in the end when it wins over the BiLSTM-CRF tool.

**(iii) High Masking (15-20%)** As masking increases to higher levels, all tools experience a substantial increase in their deidentification performance. ClinicalBERT continues to outperform both BiLSTM-CRF and PHILTER in this range. BiLSTM-CRF and PHILTER see major improvements in the performance, and Philter continues to outperform BiLSTM-CRF.

4.2 Error Analysis

4.2.1 Incorrect Mappings with Low Masking

Some EHR notes were incorrectly mapped to headers despite having low masking. In the results where low-masked EHR notes were mapped to headers, around 10-30% of the mappings were incorrect, depending on the amount of masking and the tool used. As expected, this percentage generally increased with more masking, as more PHIs and quasi-identifiers were removed, making it harder to map the notes correctly.

Surprisingly, about half of the incorrectly mapped low-masked EHR notes contained highly identifiable PHIs, such as addresses, MRN numbers, and birthdates which should in practice make the mapping easy. Additionally, the probability of these incorrect mappings was unexpectedly high. These two factors highlight areas where the model needs improvement. Shown in Table 3 are examples of such EHR notes. Please note that while we have retained the kind of identifiers as found by us, we have morphed their values for the sake of privacy.

**Table 3.** Table showing low-masked but incorrectly mapped (morphed) EHR notes, with corresponding mapping probabilities. Even though only a few identifiers were masked during deidentification and many remained unmasked (2nd column), the reidentification model failed to correctly map the note to the corresponding note header and, in fact, incorrectly mapped it with high confidence (3rd column).

| Note | Profile | Probability |
|---|---|---|
| The Florida Hospital - Harvard Cryogenic Center Department Surgical Pathology Laboratory Monn 2412 Inquiry:(421)321-5313 421 West 88th Street, FL,FL 14213 Fax:(531)132-5313 Surgical Pathology Report S05-241235 Specimen Date: ********** In Lab | 'city': MIAMI 'state': 'FL' | 0.95 |
| Date: ********* 09:30 Report Date: ********* Patient: WOOD, MARK Submitted by:JOHN E. FERRARI, M.S. 4213 11TH STREET Cryogenic Surgery, Box 123 MIAMI, FL 19421 983 West 88th Street, R-992 FL 95213 Telephone: (885) 823-5213 Fax: 1(421) 821-2914A FLH-MRN: 92819291 Encounter #: 821728492 Location: W33E Age/Sex: 43 / M CLINICAL INFORMATION: *[68 tokens omitted…]* | 'mrn': '92819291' 'zip': '19421' | |
| The Florida Hospital - Harvard Cryogenic Center Department of Pathology Surgical Pathology Laboratory Moonn 1521 Inquiry:(421)531-2132 524 West 48th Street, FL,FL 24123 Fax:(553)763-2342 Surgical Pathology Report T31-94312 Specimen Date: ********** In Lab Date: ********* 13:13 Report Date: ********** Patient: JOHN, WICK Submitted by:****** Te, M.D. 251 EAST 12TH STREET WMC ***** Cryology APT 13B 521 West 18th Street, Y-341 Florida, FL 95123 Telephone: (912) 421-4951 Fax: 1(953) 912-2912 FLH-MRN: 92142123 Encounter #: 95123124 Location: CLCC Age/Sex: 63 / M CLINICAL INFORMATION: *[91 tokens omitted…]* | 'address_2': 'APT 13B' 'city': 'FLORIDA' 'state': 'FL' 'mrn': '92142123' 'zip': '24123' | 0.99 |
| San Francisco-Podiatrist/Harvard San Francisco-Podiatrist 512 W. 28th Street Harvard Hospital San Francisco, CA 92412 Medical College 824-853-8291 Cardiac Catheterization Report Patient Name: SRILA PRABHUPADA Date of Study: 5/10/2010 5:31:21 AM Med Rec #: 4912412 Accession #: ADS_94212 DOB/Age: 2/1/1936 - 88 years Pt. Gender: M Room: Lab 2 Patient Status: Inpatient / Urgent Referring Physician: ********** MD Path Lab Personnel: **************** MD Interventional Cryologist Jerry ***** MD Interventional Fellow Caro Mountaindes RCIS Angioplasty Specialist INDICATIONS: 512.0 Nonrheumatic aortic (valve) stenosis; 111.9 Chronic ischemic heart disease, unspecified; 951.21 Acute systolic (congestive) heart failure; 512.9 Cryomyopathy, Unspecified *[103 tokens omitted…]* | 'Birth or Death date': '2/1/1936' 'note_datetime': '3/11/2009' 'city': 'San Francisco' 'state': 'CA' 'mrn': '4912412' | 0.99 |

### 4.2.2 Correct Mappings with High Masking

Similarly, in the cases when EHR notes were highly masked, about 10-20% of the total test data falls into this category. These results highlight the capability of the reidentification model to correctly reidentify individuals even with significant masking. Upon further analysis, it was found that about half of these correct mappings were due to the presence of quasi-identifiers in the EHR notes. In one example, our system was able to reidentify an individual simply by their age, gender, and city of origin, given that no other patients in our database matched all three criteria.

### 6. Discussion

### 6.1 Additional uses of DIRI

**Auditing deidentified datasets.** One use of our tool is benchmarking and improving deidentification tools on existing datasets where demographic information is available. A main use of our method is evaluating the privacy leakage of existing datasets: before data-sharing, practitioners can train our reidentification models on their proprietary datasets to measure dataset-level leakage (as measured by reidentification top-k accuracy) as well as identifying at the sample level which deidentified notes leak the most information.

**Masking threshold-setting.** A common practical problem when performing deidentification is "where to draw the line", i.e. at what level of confidence data should be masked before exporting the dedentified dataset. In our experiments, we showed that even in the most private setting, where over 20% of the text was masked in order of decreasing token confidence as measured by our ClinicalBERT tool, we could still identify over 9% of individuals from masked data. As an example, if <1% of the data is allowed to be correctly reidentified, this indicates that with our dataset, the correct amount of token masking would be greater than 20%. An easy way to detect this threshold would be to train reidentification models at various intervals of masking and set the threshold to be the minimum amount of masking that yields reidentification below 1%. In compute-constrained scenarios, a more efficient way to determine this threshold would be to sequentially train reidentification models using binary search to find the ideal masking threshold.

## 6.2 Limitations

**Knowledge of a profile.** The most glaring limitation of our method is its reliance on patient-level metadata for each patient mentioned in a deidentified note. In other words, a patient's note cannot be statistically reidentified using our method unless demographic information about that patient is known. Although demographic data is commonly stored and easily accessible in medicine, this requirement may make it difficult to adapt our method to a broader set of domains. This is also a clear avenue for future improvement: by integrating publicly-available data sources external to the EHR, we could improve reidentification and quasi-identifier detection performance.

**Profile incompleteness.** When patient-level demographic information is known, our deidentification algorithm is only as good as the data used for reidentification. Our models are not "omniscient" in the sense that they can only reidentify a note based on an included piece of PHI *if that key is also present in the demographic information*. As an example, if a social security number is present in a deidentified clinical note, our model will only be able to correctly reidentify that patient if the patient's SSN is also stored as demographic information in the structured database.

**Necessity of a training set.** Our method customizes itself to each specific deidentification technique using a training set of separate notes and patients deidentified using the same technique. In this way, our method is naturally adversarial, and can pick up on subtle PHI leaks that would not be detected by a general-purpose tool. However, the dependence of our system on tens of thousands of additional notes for training is a limitation that we could hope to resolve in future work.

**Tool inconsistencies.** We also exhibit a few differences between the deidentification tools that may affect our analysis. Both the neural network-based tools (ClinicalBERT and the LSTM) output a confidence score for masked PHI, so we are able to interpolate their masking rate intelligently, and only mask the most important PHI at low levels of masking. As Philter does not compute confidence scores, we instead chose to randomly mask the set of tokens identified by Philter in a random order, which may affect its performance, especially at low rates of masking. Finally, the CRF tool was mainly evaluated on Dutch data, which may explain its general lack of effectiveness at deidentifying our medical notes, which are all written in English. Additionally, none of the deidentification tools have been trained with a hierarchical understanding of PII importance. Uniquely identifiable information such as names and MRN numbers carries a higher risk of re-identification compared to more general PIIs like city names.

## 6.3 Future work

**Larger and more powerful reidentification models.** Our models are on the order of hundreds of millions of parameters, which are efficient to train and run in a biencoder setup but hundreds or thousands of times smaller than the largest, most intelligent models used for some reasoning-intensive medical tasks. Our models run in seconds on a single consumer-grade GPU. Due to the utmost importance of privacy in the medical domain, it is not unreasonable to expect future systems might be willing to spend more time, money, and computing to run much larger models to perform this reidentification task.

**A universal reidentification model.** Our reidentification system relies on tool-specific, institution-specific datasets to "train" and run in order to evaluate a potentially-deidentified dataset. If we were able to gather data from a large number of deidentification tools and domains we could train a large, universal reidentification model that could be used in the future to reidentify patients without a training set.

**Hiding in plain sight.** One promising avenue for deidentification is the idea of *hiding in plain sight*[23], where some PHI is randomly replaced with synthetic PHI. Our system could naturally be integrated into reidentifying datasets redacted using these techniques. We leave it to future research to evaluate DIRI on datasets deidentified via hiding in plain sight.

**Going beyond k-anonymity.** Our method depends on the principle that if a patient is not the top-1 reidentified individual from a clinical note, that note is deidentified. This definition relates to the privacy principle of k-anonymity, which is intuitive but not the only existing definition of privacy. It may have some undesirable principles, like scaling with the size of the patient dataset, and ignoring the fact that if a patient is the *second* most-relevant to a clinical note, this may constitute a privacy violation in certain scenarios. For example, a parent and child may have similar names and addresses; if information is redacted from a note so that our system misidentifies the note as belonging to an individuals parent, should we consider this note deidentified? We leave it to future work to explore applications of other definitions of privacy in reidentification models.

# 7. Conclusion

The duality between deidentification and reidentification reveals an exciting new path forward for the field of text anonymization. Along with enabling comparison between existing deidentification tools' performance, our tool reveals a way for institutions to set masking thresholds for existing tools, and detect leakage in existing anonymized datasets. These findings show that while existing popular tools may achieve high scores on deidentification benchmarks, they still may leak some information that can lead to reidentification by our powerful neural models. Hopefully, this study will inspire improvements on both sides and help researchers jointly improve systems for both deidentification and reidentification.